\documentclass{llncs}
\usepackage{graphicx}
\usepackage{xspace}
\usepackage{qtree}

  \usepackage{soul, color}
  \sethlcolor{green}
  \usepackage{url}
  
  \usepackage[ruled,vlined,noresetcount]{algorithm2e}

\usepackage{lscape}
\usepackage{framed}
\usepackage{url}
\usepackage{amsmath}
\usepackage{color, colortbl}
\usepackage[table]{xcolor}
  \usepackage{soul, color}
  \sethlcolor{green}

\usepackage{multirow}

 \usepackage[neveradjust]{paralist}
\usepackage{tipa}

\DefineNamedColor{named}{Blue}          {cmyk}{1,1,0,0}
\DefineNamedColor{named}{BrickRed}      {cmyk}{0,0.89,0.94,0.28}
\DefineNamedColor{named}{Brown}         {cmyk}{0,0.81,1,0.60}
\DefineNamedColor{named}{ForestGreen}   {cmyk}{0.91,0,0.88,0.12}
\DefineNamedColor{named}{Purple}        {cmyk}{0.45,0.86,0,0}
\DefineNamedColor{named}{Red}           {cmyk}{0,1,1,0}
\DefineNamedColor{named}{Black}         {cmyk}{0,0,0,1}
\DefineNamedColor{named}{RoyalBlue}     {cmyk}{1,0.50,0,0}

\begin{document}

\title{Grammar rules for the isiZulu complex verb
}


\author{C. Maria Keet\inst{1}         \and
        Langa Khumalo\inst{2} 
}


\institute{Department of Computer Science, University of Cape Town, South Africa,               
              \email{mkeet@cs.uct.ac.za}           
           \and
           Linguistics Program, School of Arts, 
           University of KwaZulu-Natal, South Africa, 
              \email{khumalol@ukzn.ac.za} 
}


\maketitle

\begin{abstract}

The isiZulu verb is known for its morphological complexity, which is a subject for on-going linguistics research, as well as for prospects of computational use, such as controlled natural language interfaces, machine translation, and spellcheckers. To this end, we seek to answer the question as to what the precise grammar rules for the isiZulu complex verb are (and, by extension, the Bantu verb morphology). To this end, we iteratively specify the grammar as a Context Free Grammar, 
and evaluate it computationally. The grammar presented in this paper covers the subject and object concords, negation, present tense, aspect, mood, and the causative, applicative, stative, and the reciprocal verbal extensions, politeness, the wh-question modifiers, and aspect doubling, ensuring their correct order as they appear in verbs. The grammar conforms to specification.\\

\end{abstract}

\section{Introduction}
\label{intro}

While South Africa recognises eleven official languages, only English and Afrikaans have significantly invested in computational resources. IsiZulu, which is the most widely spoken language in South Africa, still remains under-resourced. In this article we focus on the development of perfect grammar rules for the isiZulu verb and by extension Bantu verb morphology. 
	It is notable that a small Definite Clause Grammar and POS tagger for isiZulu has been proposed and is available online \cite{Spiegler10}. It covers only a fraction of the complexities of the isiZulu verb; for instance, it addresses only one extension to the exclusion of the causative, applicative, and the reciprocal extensions. Other attempts to formal approaches to isiZulu morphology focus predominately on nouns rather than verbs \cite{Pretorius09fsm,Pretorius12}, or describe only a few sample regular expressions that cover a very small fraction of the verb \cite{Bosch05}. The morphology of the verb is widely regarded as the most interesting theoretically. Sections~\ref{sec:zuluIntro} and~\ref{sec:relworks} provides a brief discussion on this interesting grammatical category whose complexity presents challenges to the computation and generation of grammar rules. Traditional accounts on isiZulu grammar are based on dated sources \cite{Doke27,Doke35} and limited accounts on Wikipedia. There is no comprehensive synchronic grammar of isiZulu yet.
	
We present a morphological analysis of the isiZulu verbal extension and rules for that. This is done in order to create a spell checking and part-of-speech tagging of the verb in isiZulu. We explore the means to automate the checking of the complex verb morphology. We ultimately address the following question: 
{\em What are the precise grammar rules for the isiZulu verb (and, by extension, the Bantu verb morphology)?}
We thus formalise the grammar for the isiZulu verb as a Context-Free Grammar. This grammar is subsequently represented computationally so as to test its correctness with respect to specification, using a set of words and generating their derivations in the JFlap tool. The grammar covers not only the usual subject and object concords, but also negation, present tense, aspect, mood, and the verbal extensions such as the causative, applicative, stative and the reciprocal, politeness, the wh-questions modifiers, and aspect doubling. 

The paper is structured as follows. Section~\ref{sec:zuluIntro} gives a synchronic outline of the isiZulu verb morphology and also highlights comparative salient features that are characteristic of Bantu languages and Section~\ref{sec:relworks} discusses related works. The main contribution, the formalised account of the isiZulu verb for present tense, is presented in Section~\ref{sec:main} and evaluated in Section~\ref{sec:eval}. We 
conclude in Section~\ref{sec:concl}.



\section{Basics of the isiZulu verb}
\label{sec:zuluIntro}

IsiZulu is a Bantu language that belongs to the Nguni\footnote{A term used by Guthrie \cite{guthrie1971comparative} to classify isiZulu, isiXhosa, isiNdebele and siSwati in Group S, Zone 40.}  group of languages.  It has close affinity to other Nguni language varieties. Bantu languages have a characteristically agglutinating morphology, which makes their structure rich and complex. The agglutinating typology is not unique to Bantu languages as other agglutinating languages with extremely complex morphology include Turkish, Hungarian, and Finnish \cite{Durrant13}. In characterising the complexity of the verbal constructions in Bantu languages, \cite{Wald87} (p291) states that the morphology of the verb shows ``[...] the fullest extent of the agglutinative nature of the Bantu language family''.  Such complex morphology presents a lot of challenges in attempts to develop computational technologies in isiZulu. 

The isiZulu verbal morphology typically comprise of a verb root (VR) to which extensions such as the causative, applicative, reciprocal, passive etc. are suffixed and to which morphemes that encode negation (NEG), subject marker (SM) and object marker (OM) that cross-reference noun phrases (NPs), tense/aspect, modality, etc. are prefixed. 

At the core of the verbal structure is a root morpheme, which is called the verb root (VR). The VR forms the nucleus of the verbal morphology. This core element supports a number of affixes, both prefixes and suffixes. Each affix type occupies a specific position in the verbal morphology. The affixes include the SM, the OM, Tense Aspect and Mood (TAM), and various derivational extensions. The verb is characteristically terminated with a final vowel (FV) and this final vowel of the verb may encode mood, tense, polarity and potential modality. Fig.~\ref{fig:verb2} illustrates the complex verb in Bantu.

 \begin{figure}[h]
\centering
\Tree [.{Verb} 
		{NEG}  [.{I$^{\prime\prime}$} 
			{SC} [.{I\1} 
				{T/A} [.{M\1} 
					{MOD}  [.{Macro-Stem} 
						{OC} [.{Verb stem} 
								[ {Verb Root} [.{\textit{Ext$_{\mbox{o}}$}}
												{\textit{C}} {\textit{A}} {\textit{R}} {\textit{P}}
											] ].{Verb Rad} [{\textit{Post-final}} ].{\textit{Final Vowel}} 
													]
						] 
				 	] 
			 	] 
			] 
		] 
	
    \caption{The structure of a complex verb in Bantu, where the elements not in italics font are considered to be the canonical verb structure. NEG: negative; SC: subject concord; T/A: tense/aspect; MOD: mood; OC: object concord; Verb Rad: verb radical; C: causative; A: applicative; R: reciprocal; P: passive.}
    \label{fig:verb2}
\end{figure}

Khumalo \cite{Khumalo07} (p79) proposes a verb slot system for the complex verbal form\footnote{The following abbreviations are used: A=aspect; ADV=adverb; APPL=applicative; CONT=continuous tense; EXCL=exclusive aspect; Ext=extension; FV=final vowel; M=mood; NEG=negative tense; OC=object concord; PROG=progressive tense; Rad=radical; SG=singular; SC=subject concord; T=tense; VR=verb root; VS=verb stem}  in Ndebele, which is applicable to isiZulu, as included in Table~\ref{tab:slots}.
\begin{table}[h]
\caption{Bantu verb slot template, adapted from \cite{Meeussen67}; $\sim$: also realised as}
\begin{center}
\begin{tabular}{|p{1.4cm}|p{1.0cm}|p{1.0cm}|p{1.2cm}|p{1.2cm}|p{1.0cm}|p{1.7cm}|p{0.9cm}|p{1.7cm}|}
\hline
{\bf Slot} & {\bf Pre-initial} & {\bf Initial} & {\bf Post-initial} & {\bf Pre-radical} & {\bf Radi- cal} & {\bf Pref-final} & {\bf Final} & {\bf Post-final} \\  \hline   \hline           
{\em Function} & TAM, NEG, clause type & SM & TAM, NEG, SM & OM & VR & TAM, valence change (CARP) & FV & Participant, NEG, clause type  \\  \hline
{\em Example}  & a & ngi & za, nga & ba & khal & is (el; an; w) & a & (ni $\sim$ nini)$^1$  \\  \hline
\end{tabular}
\end{center}
\vspace{-2mm}
$^1$ The plural suffixes denote both general plurality and honorific plurality.
\label{tab:slots}
\end{table}%
%
%
The prefixed morphemes differ from suffixed extensions in both form and function. Formally the suffixes have a -VC- structure, as opposed to the regular CV syllable structure. Functionally the verbal extensions affect the argument structure \cite{Mchombo07} (p203). Example (1) shows the morphological organisation of the verb in isiZulu.

$
\begin{array}{llllllllll}
\mbox{(1)} & \mbox{\em Aba-shana }  & \mbox{\em ba-ya-zi-theng-is-el-an-a}  &   \mbox{\em izimpahla} \\
			& \mbox{2.Children}  &   \mbox{2SM-Pres-8OM-buy$_{\mbox{{\sc vr}}}$-CAUS-APPL-REC-FV}  &  \mbox{8.clothes} \\
& \mbox{`The children} & \mbox{are selling the clothes to each other'}
\end{array}
$

The VR {\em -theng-} `buy' supports the extensions {\em -is-} for the causative, {\em -el-} for the applicative, {\em -an-} for the reciprocal, and the prefix clitics {\em ba-} for the `subject marker', 
{\em -ya-} for the `tense', and {\em -zi-} for the `object marker'. 

The verb extensions interact in complex ways with the valency of the base verb. The extensions for several languages are listed in Table~\ref{tab:vext}. Semantically (with the exception of the passive extension) they alter the number of participants expressed by the verb. Grammatically they alter the number of arguments present expressed by an NP or a pronominal element. 

\begin{table}[h]
\caption{Verbal extensions in Proto Bantu, Swahili, isiZulu and isiNdebele. *: morphemes belong to an ancestor language or proto form, \textipa{V}: precise phonetic form of the vowel. (Adapted from \cite{Schadeberg03}, p72.)}
\begin{center}
\begin{tabular}{|p{4.7cm}|l|l|l|l|}
\hline
{\bf Derivational Extension} & {\bf Proto Bantu} & {\bf Swahili} & {\bf Zulu} & {\bf Ndebele} \\  \hline   \hline           
causative & *-i-/-ici- & -ish-, -esh- & -is- &-is-  \\  \hline
applicative/dative & *-il- & -i-, -e- & -el- & -el- \\  \hline
reciprocal/associative & *-an- & -an- & -an- & -an- \\  \hline
passive & *-\textipa{V}-/-ib\textipa{V} & -w- & -iw-, -w- & -iw-, -w- \\  \hline
stative/neutro-passive/positional & *-am- & -ik-, -ek- & -ek- & -ek-, -akal- \\  \hline
reversive/separative & *-\textipa{V}l-; \textipa{V}k- & -u-, -o- & -ul- & -ul-, -ulul-, -uk- \\  \hline
neuter & *-ik- & & -akal- &  \\  \hline
extensive & *-al- & & & -isis- \\  \hline
repetitive & *-ag- $\sim$ -ang- & & &  \\  \hline
impositive & *-ik- & & &  \\  \hline
tentive/contative & *-at- & & &  \\  \hline
\end{tabular}
\end{center}
\label{tab:vext}
\end{table}%

%
%
%


\subsection{Concordial agreement}

The term agreement in Bantu is often used alongside the term concord. These two terms are sometimes used interchangeably. Agreement occurs when grammatical information appears on a verb which typically is not the source of that information. This is done through a series of agreement markers called concords that are affixed to the verb. The noun or pronoun is said to govern the agreement of all words associated with it in a syntactical relationship \cite{Zawawi79} (p8). Agreement is thus a cross-referencing device for subjects and objects. Table~\ref{tab:concords} shows the conjugation of the verb in isiZulu for all the noun classes and persons. As shown in the table the verb not only takes the subject and object concords, but also the negative subject concord.

%

As stated above, the verbal structure consists entirely of bound morphemes. These are the VR and a number of affixes such as the subject concord (SC), the object concord (OC), Tense Aspect Mood (TAM), and various other derivations (CARP), typically terminated with a final vowel (FV). The example below is a Chishona verb {\em ndichaenda} `I will go'

$
\begin{array}{llllllllll}
\mbox{(V2)} & \mbox{\em ndi}  & \mbox{-} &  \mbox{\em cha}  & \mbox{-} &  \mbox{\em end} & \mbox{-} &  \mbox{\em a}	&& \mbox{{\em Chishona: ndichaenda}}\\
& \mbox{1.SC}  & \mbox{-}  &  \mbox{1.TM}  & \mbox{-} &  \mbox{Root} & \mbox{-} &  \mbox{FV}	&& \\
& \mbox{`I'}  &&  \mbox{`will'}  && \mbox{go}  &&  && \mbox{{\em `I will go'}}\\
\end{array}
$

%

%

\begin{table}[t]
\caption{Basic verb conjugation. NC: noun class; SC: subject concord; OC: object concord; NEG SC: negative subject concord.}
\begin{center}
\begin{tabular}{|p{0.7cm}|p{0.8cm}|p{1.5cm}|p{0.8cm}||p{1.9cm}|p{0.8cm}|p{1.5cm}|p{0.8cm}|}
\hline
\multicolumn{4}{|c||}{\bf Conjugation for noun classes} & \multicolumn{4}{|c|}{\bf Conjugation for persons}  \\ 
{\bf NC} & {\bf SC} & {\bf NEG SC} & {\bf OC} & {\bf Pers. Pron.} & {\bf SC} & {\bf NEG SC} & {\bf OC} \\ \hline   \hline           
1 & u- & aka- & -m- & I & ngi- & angi- & -ngi- \\ \cline{5-8}
 2 & ba- & aba- & -ba- & you (sing.) & u-  & awu- & -ku- \\ \hline
  1a & u- & aka- & -m- & he/she & u- & awu- & -m- \\ \cline{5-8}
   2a & ba- & aba- & -ba- & we & si- & asi- & -si- \\ \hline
     3a & u & aka- & wu & you (pl.) & ni- & ani- & -ni- \\ \cline{5-8}
      (2a) & ba- & aba- & -ba- & they & ba- & aba- & -ba- \\ \hline
     3 & u- & awu- & -wu- &  &  & & \\
     4 & i- & ayi- & -yi- &  & & &\\ \hline
      5 & li- & ali- & -li- & &  & &\\
       6 & a- & awa- & -wa- & &  & &\\ \hline
        7 & si- & asi- & -si- &  & & &\\
         8 & zi- & azi- & -zi- & & & &\\ \hline
         9a & i & ayi- & yi &  & & &\\ 
          (6) & a- & awa- & -wa- &  & & &\\ \hline                  
          9 & i-& ayi- & -yi- &  & & &\\
           10 & zi-& azi- & -zi- &  & & &\\ \hline
            11 & lu- & alu- & -lu- &  & & &\\
             (10) & zi- & azi- & -zi- &  & & &\\ \hline
              14 & bu- & abu- & -bu- &  & & &\\ 
               15 & ku- & aku- & -ku- &  & & &\\ \hline
\end{tabular}
\end{center}
\label{tab:concords}
\end{table}

\subsection{More on clitics of the verb}


The elements prefixed to the verb stem in isiZulu are usually referred to as clitics. Clitics are independent syntactic elements which appear as part of the host word. This independent element is involved in a morphological merger to appear phonologically as part of a derived word. Example 3 is illustrative:
$
\begin{array}{lllllllllll}
\mbox{(V3)} & \mbox{\em ngi}  & \mbox{-}&  \mbox{\em m}  & \mbox{-}&  \mbox{\em bon} & \mbox{-}&  \mbox{\em a} & \mbox{\em kusasa} 	&& \mbox{{\em Ngimbona kusasa.}}\\
& \mbox{1.SC}  & \mbox{-} &  \mbox{1.OC}  & \mbox{-}&  \mbox{Root} & \mbox{-}&  \mbox{FV}	&& &\\
& \mbox{`I'} &&  \mbox{`him'}   &&  \mbox{see}  && & \mbox{tomorrow}  && \mbox{{\em `I (will) see him tomorrow.'}}\\
\end{array}
$

\noindent The independent elements {\em ngi-} and {\em m-} merge to form a derived word {\em ngimbona}. The clitics are thus syntactic elements, which lack phonological independence. They cannot stand or appear on their own. It is clear that syntactically they are words but phonologically they are not. They are not viewed as phonological words because they fail to satisfy the minimality condition for being a word in Bantu. The Bantu condition is that a word has to minimally consist of two syllables. In the example above, {\em ngi-} is a single syllable and {\em m-} is also a single syllable known as ``syllabic {\em m}''. The notion of clitics and their grammatical status in Bantu is still a very interesting one. 

The isiZulu verbal suffixes are also bound morphemes without any independent status, hence they are also clitics. They are involved in the determination of expressible NP arguments within the sentence. As stated earlier, these include the morphology for encoding the causative, applicative, reciprocal, passive, stative, etc. These suffixes, together with the VR, are terminated by the FV {\em -a} and together make up the verb stem (VS) as shown in Figure~\ref{fig:verb2}. The following example shows the VR plus the verbal extensions.

$
\begin{array}{lllllllll}
\mbox{(V4)} &\mbox{\em bon}  & \mbox{-} & & &   \mbox{\em a} & \mbox{{\em bona}} & \mbox{`see'} & \mbox{un-extended verb}\\
& \mbox{VR}   & \mbox{-} & & &  \mbox{FV} & & & \\
& \mbox{\em bon}  & \mbox{-}& \mbox{\em\textbf{is}} &\mbox{-}& \mbox{\em a}  & \mbox{{\em bonisa}} & \mbox{`make see'} &  \mbox{extended verb}\\
&   \mbox{\em bon}  &\mbox{-}& \mbox{\em\textbf{el}} &\mbox{-}& \mbox{\em a}  & \mbox{{\em bonela}}  & \mbox{`see for'} & \mbox{extended verb} \\
&  \mbox{\em bon}  &\mbox{-}& \mbox{\em\textbf{an}} &\mbox{-}& \mbox{\em a}  & \mbox{{\em bonana}}  & \mbox{`see each other'} & \mbox{extended verb} \\   
\end{array}
$


\noindent While the suffixes (or verb extensions) clearly introduce a new syntactic element, they however are themselves not independent. They cannot stand as phonological words on their own, hence, they are clitics. 
%
The clitics in Bantu can co-occur with the verbal extensions. However, when this happens, they are attached outside the final vowel. 
The extensions appear to be more intimately connected to the host VR. Crucially, while the VS is the domain of a number of linguistic processes, its influence is not extended to the suffixed clitics. It is thus assumed that the VS has lexical integrity. 
This makes the VS an important subdomain in the morphological structure of the verb. The VS is thus the domain of lexical processes in Bantu \cite{Mchombo04}.

\subsection{Aspects of isiZulu Tense}

Bantu languages typically consist of rich tense and aspect systems, characterised by various temporal distinctions \cite{Lindfors03}. The complexity of grammaticalised tense and aspect in isiZulu is exemplified by its five tenses. The tenses include the remote past, recent past, present, immediate future and remote future tense. The three aspectual forms are the simple, progressive and exclusive aspect.  

IsiZulu makes productive use of its grammatical aspect system. The Progressive aspect in isiZulu is denoted by the affix {\em -sa-}. Whilst conveying an ongoing action/state/event, the morpheme also carries an inherent adverbial meaning of `still' as shown in the example below. 
\begin{description}
\item \hspace{3mm} Ngi-{\bf sa}-fund-a \hspace{8mm} 	isiZulu\\
	1SG-PROG-VR-FV  isiZulu\\
	`Even now I am still studying Zulu'
\end{description}

There is no direct adverb (lexical item) for the English word `still' in isiZulu. Instead it is expressed using the adverb {\em namanje}, which directly translated means `even now'. The rich expression of temporal events and situations in isiZulu, is further highlighted in the following example.

\begin{description}
\item \hspace{3mm} Na-manje  ngi-{\bf sa}-fund-a    \hspace{9mm}      isiZulu\\
	Cl-ADV  \hspace{1mm}   1SG-PROG-VR-FV isiZulu\\
	`I am still studying isiZulu'
\end{description}

Similarly, the Exclusive morpheme {\em se-} expresses an inherent adverbial aspect, meaning `now'. This morpheme may be used with the adverb {\em manje} `now', thereby expressing double aspect comprising of the grammatical aspect ({\em se-}, now) + grammatical aspect ({\em manje}, `now'). This phenomenon has been referred to as aspect doubling, and is illustrated below. 

\begin{description}
\item \hspace{3mm} Se-ngi-ya-fund-a    \hspace{15mm}      manje\\
	EXCL-1SG-CONT-VR-FV  ADV\\
	`Now, I am now studying'
\end{description}

The productive nature of Exclusive and Progressive aspect morphemes in isiZulu has not received considerable attention. The Exclusive morpheme {\em se-} may be used with adverbial structures conveying similar meanings in isiZulu, while this is proscribed in the English language. 

Section~\ref{sec:zuluIntro} has thus shown the complexity of the morphology of the verb. It has shown that not only does isiZulu verb get inflected before the VR but also after the VR through a whole gamut of clitics that have an effect on the construction of a whole sentence. This is not unique to isiZulu but is characteristic of other Bantu languages like Chishona. It is thus this complexity of verbal morphology which presents challenges in the development of computational technologies in isiZulu.




\section{Related work on isiZulu verbs}
\label{sec:relworks}

The verb in Bantu has received considerable attention (cf. \cite{Hyman91,Mabugu01,Mchombo04} etc.).   This is because it is arguably the most interesting grammatical category in linguistic theory. Many accounts in Bantu have sought to explicate the many salient morphosyntactic properties of the verb using different generative theoretical approaches. Buell \cite{Buell05}  is the most recent comprehensive study of isiZulu verb. Buell discusses the isiZulu verb using a restrictive theory of syntax, which is premised on the assumption that there is a close relation between the morphology and the syntax. His account covers an array of inflectional elements such as mood, sub-mood, and polarity, subject and object agreement. Buell also makes reference to, albeit briefly, the verbal suffixes such as the applicative \cite{Buell05}. In a study such as his, it is impossible to be exhaustive. In this study, however, we cover the causative, applicative, reciprocal and the passive. Earlier studies on the Zulu verb are  \cite{Beuchat66}, whose study focuses on the verb and its conjugation of various subject concords and their allomorphs, tense and mood conjugational morphemes. Beuchat \cite{Beuchat66} does not make reference to derivational extensions. 

As we seek to have a precise, formal, representation of the isiZulu verb, we also consider computational processing of the verbs, for they require a structured representation to work computationally.
Regarding controlled natural languages and natural language generation, there are only two recent papers \cite{KeetK14rule,KeetK14cnl}, which cover verbs only to the extent of noun class-appropriate singular present tense when verbalising simple existential quantification involving object properties.
Some literature on computational linguistics for isiZulu exists that is relevant to some extent, being morphological analysers. Among these works, the Ukwabelana corpus and related materials \cite{Spiegler10} is most comprehensive and is the only one with online source material. Besides the corpus and limited semi-automated POStagging, Spiegler et al. developed a basic Definite Clause Grammar (DCG)\footnote{Available from: \url{http://www.cs.bris.ac.uk/Research/MachineLearning/Morphology/resources.jsp}; last accessed on August 19, 2015.}, of which a relevant section is shown in Fig.~\ref{fig:dcg}. 
\begin{figure}
\begin{verbatim}
v --> neg, spfn, asp, opf, vr1, vsf_neg.
v --> neg, spfn, asp, vr1, vsf_neg.
...
v --> spfi, asp, opf, vr1, vsf.
v --> spfi, asp, vr1, vsf.
...
v --> spfp, vr1, vsf.
...
v --> spfs, opf, vr1, vs.
...
vr1 --> vr, xa.
vr1 --> vr, xc.
vr1 --> vr, xn.
vr1 --> vr, xp.
vr1 --> vr, xr.
vr1 --> vr.
\end{verbatim}
\caption{Selection of DCG statements from the online supplementary material to  \cite{Spiegler10} (``...'' means line(s) omitted here).}
\label{fig:dcg}
\end{figure}
The first to note is that while it has each of the ``CARP'' ({\tt xc} etc.; bottom part), it has only ever one of them. This constitutes a subset of the possibilities, as multiple ones can be appended and as they appear in a certain order. Also, the passive ({\tt xp} in the CFG above), which causes changes in the concords in the verb, is not catered for, nor are the politeness prefixes ({\em aw-}, a.o.) and tenses other than present tense, nor imperative. That is, it covers a subset. That said, it is already useful and at least it can be extended, unlike related works such as \cite{Bosch05,Pretorius09fsm,Pretorius12}. Bosch and Eiselen \cite{Bosch05} report on a basic spelling checker that  is based on a set of regular expressions. They illustrate 4 examples that show a few permutations for a verb, e.g., 
\begin{verbatim}
 /^(ba)(ya)?(ngi)?(.+)(el)?(a|c)(ni|phi)?$/
 \end{verbatim}
 \vspace{-3ex}
which is a subset of the conjugation ({\em ba-} for 3rd person plural) and CARP ({\em -el-}) and no details of its implementation is provided \cite{Bosch05}.  
A related work on morphological rules focuses on nouns 
\cite{Pretorius09fsm}.
 The bootstrapping approach presented in \cite{Pretorius12} considers the copulative (and a few other word categories) but not verbs in general. 
 Assuming 
 that the lexc and xfst rules as described in \cite{Pretorius03fsm} do exist, then its coverage of verb features is incomplete, notably missing mood and aspect, applicative, reciprocal, stative, politeness, and wh-ending. While their approach of figuring out which CARP extensions are permitted with a verb root is interesting (relying on the noun forms), it results in rules that are too restrictive: ``by explicitly listing the noun stems of the verb root -fund- no suffixes other than -a, -el-o, -i, -is-an-o, -is-i, -is-o, -is-wa, and -o will occur with -fund-.'' (emphases omitted) \cite{Pretorius03fsm}, but words such as {\it awufunde} `[could we/you] please study' and {\it usafundaphi} `where are you [still] studying?' are valid verb forms.

Concerning verbs in other Bantu languages, several rules for Setswana (also an official language in South Africa) verbs have been implemented in xfst \cite{PretoriousR09}, but it is not clear how much of the grammar of the verb was covered. Further afield from the languages in South Africa, there are exploratory results for Ekegusii (a Bantu language spoken in Western Kenya) with several regular expressions in xfst zooming in on the difficulties of tone in relation to verbs \cite{Elwell05}, and there is a systematic account of the Runyakitara (a Bantu language spoken in Uganda) verb implemented in fsm2, including both the grammar and context-dependent rewriting rules that handle morpho-phonological and orthographical issues \cite{Katushemererwe10}. 
 
From a scientific methodological viewpoint, there is no clear `winner' between the data-oriented approach and the knowledge and rules-based approach to obtain the grammar; or the empirical and the rational paradigms. The data-based techniques, notably machine learning \cite{Spiegler10,Getao00}, have the hurdle of finding or creating a representative enough corpus and at least some rules to process them, whereas the rules-based techniques face the issue of a dearth of up-to-date, structured, grammar books, having to start afresh with formalising the grammar as grammar or regular expressions. Our literature survey indicates the latter approach is used considerably more often for Bantu languages \cite{Pretorius09fsm,Elwell05,Katushemererwe10,Pretorius03fsm,PretoriousR09}. However, use/preference does not imply more effective. 

\section{Structured representation of the isiZulu verb}
\label{sec:main}

Methodologically, theoretically, and technically, there are multiple ways of specifying the grammar of a POS category; e.g., using a grammar such as a DCG, regular expressions, or their more abstract representation with an automaton (PDA for a CFG). While for the small subset of prefixes for noun classes and some simple verb forms it certainly is easier to design an NFA, transform it into a DFA and from there into a RE, there are so many options with the verbs that the automaton would become too large and wieldy. Moreover, the cross-dependencies of elements before and after the verb root indicates that a regular expression is not expressive enough and may need a CFG rather than an RG. To create the structured representation of the isiZulu verb that is computationally useful, we build it up stepwise from a linguistic pattern, to some quasi regular expressions  that in turn revealed a pattern, and from there to a basic grammar, which in turn was extended with other verb features. For reason of exposing this incremental methodological approach to the design of the grammar, we report on the component-steps of one cycle, and subsequently only the outcome of the subsequent cycles, which amount to extensions of the grammar obtained in the first round. The additions to the first cycle were---and can be---done in arbitrary order. 

\subsection{First iteration}
From the general linguistic structure of the isiZulu verb as depicted in Fig.~\ref{fig:verb2}, we obtain the full set of `slots' of the verb's basic components as follows:
\begin{description}
\item[R0:] [NEG] [SC] [T/A] [MOD] [OC] [VR] [C] [A] [R] [P] [FV]
\end{description}
with [VR] being the verb root at the centre. Each NEG, SC etc. has its own set of characters for each noun class; see Table~\ref{tab:concords}. For the CARP, we have, as a general rule, C = {\em is}, A = {\em el}, R = {\em an}, and for P = {\em w}, though there is some phonological conditioning for A and P. 


\subsubsection{First part before the VR}
\label{sec:prefix}

Lets consider first what comes before the verb root (VR), with the subject present and active, and both in the positive (thus FV={\em a}) and in the negative (FV={\em i}), and assuming there is an object after the verb, so that OC can be omitted (see below for OC inclusion). Then the following patterns are permissible (italicised):
\begin{description}
\item[--]  {\em [SC]} [VR] [FV=a]
\item[--]  {\em [SC] [MOD]}  [VR] [FV=a]
\item[--]  {\em [SC] [T/A] [MOD]} [VR] [FV=a]
\item[--] {\em [NEG] [SC]}  [VR] [FV=i]
\item[--] {\em [NEG] [SC] [MOD]}  [VR] [FV=i]
\item[--] {\em [NEG] [SC] [T/A] [MOD]} [VR] [FV=i]
\end{description}
This can be captured by the following two quasi regular expressions (where the NEG, SC, T/A, MOD, and VR are to be replaced by the actual strings):
\begin{description}
\item[R1:]  {\tt [SC][T/A]$^{0..1}$[MOD]$^{0..1}$[VR]a}
\item[R2:]  {\tt [NEG][SC][T/A]$^{0..1}$[MOD]$^{0..1}$[VR]i}
\end{description}
Or, if the software to implement it allows for REs+rules, then:
\begin{description}
\item[R3:] {\tt [NEG]$^{0..1}$[SC][T/A]$^{0..1}$[MOD]$^{0..1}$[VR][FV]}
\item[R4:] {\bf if} NEG {\bf then} FV=i, {\bf else} FV=a
\end{description}
The OC is used if there is no explicit object named after the verb. Then we have the following options:
\begin{description}
\item[--]  {\em [SC] [OC]} [VR] [FV=a]
\item[--]  {\em [SC] [MOD]  [OC]}  [VR] [FV=a]
\item[--]  {\em [SC] [T/A] [MOD]  [OC]} [VR] [FV=a]
\item[--] {\em [NEG] [SC]  [OC]}  [VR] [FV=i]
\item[--] {\em [NEG] [SC] [MOD]  [OC]}  [VR] [FV=i]
\item[--] {\em [NEG] [SC] [T/A] [MOD] [OC]} [VR] [FV=i]
\end{description}
This amounts to the following two rules
\begin{description}
\item[R5:]  {\tt [SC][T/A]$^{0..1}$[MOD]$^{0..1}$[OC][VR]a}
\item[R6:]  {\tt [NEG][SC][T/A]$^{0..1}$[MOD]$^{0..1}$[OC][VR]i}
\end{description}
While we could combine R1, R2, R5 and R6, it then will have to go through a whole set of permutations to either check correct syntax or generate it. We currently expect it to be quicker to look ahead to the tag of the next phrase to determine whether an OC is needed, and then choose either the rules with OC or without; that is: 
\begin{description}
\item[R7:] {\bf if} next word==$\emptyset$ {\bf or} next word != noun {\bf then} use R5 or R6, {\bf else} use R1 or R2
\end{description}
where the ``next word==$\emptyset$'' essentially means that the verb is the last word in the sentence. 

\subsubsection{Second part after the VR}

The extension is added to the verb root (VR), and comes before the FV. We show a section of the rather long list of all options: 
\begin{description}
\item[--]  {\it [some prefix]} [VR] {\em [C]} [FV]
\item[--]  {\it [some prefix]} [VR] {\em [C] [A]} [FV]
\item[--]  {\it [some prefix]} [VR] {\em [C] [A] [R]} [FV]
\item[--]  {\it [some prefix]} [VR] {\em [C] [A] [P]} [FV]
\item[--]  {\it [some prefix]} [VR] {\em [C] [R]} [FV]
\item[--]  {\it [some prefix]} [VR] {\em [C] [R] [P]} [FV]
\item[--]  {\it [some prefix]} [VR] {\em [C] [P]} [FV]
\item[--]  {\it [some prefix]} [VR] {\em [C] [A] [R] [P]} [FV]
\item[--]  {\it [some prefix]} [VR] {\em [A]} [FV]
\item[--]  etc.
\end{description}
That is, the CARP stay in that order, but any one or more of them can be used, so the following quasi regular expression can be specified: 
\begin{description}
\item[R8:]  {\it [some prefix]}{\tt [VR][C]$^{0..1}$[A]$^{0..1}$[R]$^{0..1}$[P]$^{0..1}$[FV]}
\end{description}
to be implemented by filling in the actual strings in the places of the VR, C, A, R, and P, and the {\it [some prefix]} following the rules as outlined above.

\subsubsection{From quasi RE to grammar}

The quasi REs show some repetition, and especially the ``{\it [some prefix]}'' makes it look clumsy. It also can be seen there are four components: what comes before the VR, the VR, what comes after the VR, and the final vowel. This can be addressed more easily and succinctly with a generative grammar. To design that, let us first convert R1, R2, R5 and R6 into grammar notation, using the following abbreviations: v=verb (with its adornments), n=negation, s=subject concord, t=tense, asp=aspect, o=object concord, m=mood, c=causative, a=applicative, r=reciprocative, p=passive, vr=verb root, text in true type font are terminals, and spaces in the rules are not spaces in the word, but added for readability:

\%\%R1 in CFG notation

v $\rightarrow$  s vr {\tt a} $\mid$ 
			s m vr {\tt a} $\mid$ 
			s t m vr {\tt a} $\mid$ 
			s asp m vr {\tt a}

\%\%R2 in CFG notation			

v $\rightarrow$ n s vr {\tt i} $\mid$ 
			n s m vr {\tt i} $\mid$ 
			n s t m vr {\tt i} $\mid$ 
			n s asp m vr {\tt i}

\%\%R5 in CFG notation

v $\rightarrow$ s o vr {\tt a} $\mid$ 
			s m o vr {\tt a} $\mid$ 
			s t m o vr {\tt a} $\mid$ 
			s asp m o vr {\tt a}

\%\%R6 in CFG notation			

v $\rightarrow$ n s o vr {\tt i} $\mid$ 
			n s m o vr {\tt i} $\mid$ 
			n s t m o vr {\tt i} $\mid$ 
			n s asp m o vr {\tt i}		

\noindent This still will result in duplications, for these will have to be reused for CARP. To this end, we create {\sl pre} and its negated variant {\sl npre} and a {\sl post} (that can be empty, $\varepsilon$), that will surround the verb root.\\

\noindent {\sl pre $\rightarrow$ s $\mid$
				s m $\mid$ 
				s t m $\mid$  
				s asp m $\mid$ 
				s o $\mid$ 
				s m o $\mid$ 
				s t m o $\mid$ 
				s asp m o}  \\

\noindent {\sl npre $\rightarrow$ ns $\mid$
				ns m $\mid$ 
				ns t m $\mid$  
				ns asp m $\mid$ 
				ns o $\mid$ 
				ns m o $\mid$ 
				ns t m o $\mid$ 
				ns asp m o}  \\

\noindent {\sl post $\rightarrow$ c $\mid$
				c a $\mid$
				c a r $\mid$
				c a p $\mid$
				c r $\mid$
				c r p $\mid$
				c p $\mid$
				c a r p $\mid$
				a $\mid$
				a r $\mid$
				a r p $\mid$
				a p $\mid$
				r $\mid$ \newline
		\mbox{}\hspace{11mm}		r p $\mid$
				p} $\mid$
				$\varepsilon$\\

\noindent This is then put together with the verb root and final vowel:

 {\sl v} $\rightarrow$ {\sl pre vr post} {\tt a} $\mid$
						{\sl npre vr post} {\tt i}  

\noindent Let us now complete the grammar so far with the terminals.
\begin{compactenum}
\item List of subject concords and negative sc:

{\sl s} $\rightarrow$ {\tt ngi} $\mid$
				{\tt u} $\mid$
				{\tt si} $\mid$
				{\tt ni} $\mid$
				{\tt ba} $\mid$
				{\tt i} $\mid$
				{\tt li} $\mid$
				{\tt a} $\mid$
				{\tt zi} $\mid$
				{\tt lu} $\mid$
				{\tt bu} $\mid$
				{\tt ku} $\mid$ $\varepsilon$

{\sl ns} $\rightarrow$ {\tt angi} $\mid$
				{\tt awu} $\mid$
				{\tt aka} $\mid$
				{\tt ali} $\mid$
				{\tt asi} $\mid$
				{\tt ayi} $\mid$
				{\tt alu} $\mid$
				{\tt abu} $\mid$
				{\tt aku} $\mid$
				{\tt ani} $\mid$ \newline
		\mbox{}\hspace{11mm}														
				{\tt aba} $\mid$							 
				{\tt awa} $\mid$
				{\tt azi} $\mid$ $\varepsilon$


\item List of mod:

{\sl m} $\rightarrow$ {\tt a} $\mid$
				{\tt e} $\mid$
				{\tt ka} $\mid$
				{\tt ma} $\mid$
				{\tt nga} $\mid$ $\varepsilon$

\item List of tense (nothing for the simple present tense):

{\sl t} $\rightarrow$ $\varepsilon$ 

\item List of aspect (additional rules omitted in this first iteration):

{\sl asp} $\rightarrow$ {\tt sa} $\mid$ {\tt se} $\mid$ {\tt be} $\mid$ {\tt ile} $\mid$ $\varepsilon$

\item List of object concords:

{\sl o} $\rightarrow$ {\tt ngi} $\mid$
				{\tt si} $\mid$
				{\tt ku} $\mid$
				{\tt ni} $\mid$
				{\tt m} $\mid$
				{\tt ba} $\mid$
				{\tt wu} $\mid$
				{\tt yi} $\mid$
				{\tt li} $\mid$
				{\tt wa} $\mid$
				{\tt zi} $\mid$
				{\tt lu} $\mid$
				{\tt bu} $\mid$ $\varepsilon$

\item Causative:

{\sl c} $\rightarrow$ {\tt is} 


\item Applicative:

a $\rightarrow$ {\tt el}  


\item Reciprocative:

{\sl r} $\rightarrow$ {\tt an}  

\item Passive (with phonological conditioning options):

p $\rightarrow$ {\tt iw} $\mid$ {\tt w} 


\item Lexicon of verb root:

{\sl vr} $\rightarrow$ {\tt ab} $\mid$ 
				{\tt ...} $\mid$
				{\tt zwib}  

%

\end{compactenum}

\noindent This completes the first iteration: the core possibilities for present tense are 	completed with respect to R0 mentioned at the start of the section\footnote{Except that it does not take into account the swapping with OC and SC in case of P}. It can be optimised, but this is left for the implementation; here, we aimed  to be as explicit as feasible.
	

\subsection{Subsequent iterations}

The outcome of the first iteration does not fully cover all verb options. Further extensions and refinements can be made, which are introduced now in their final version, being politeness, stative verbs, wh-questions, and aspect doubling.

\paragraph{Politeness} The please and polite permissive questions have their own prefix system and a FV={\tt e}. This amounts to adding a new rule

 {\sl ppre $\rightarrow$ pl s}  

\noindent with the following terminals:
\begin{compactenum}
\setcounter{enumi}{10}
\item Please prefix, permissive prefix (none), and polite proposal doing something together, indicated with {\sl pl}:

{\sl pl} $\rightarrow$ {\tt aw} $\mid$ 
				{\tt awu} $\mid$
				{\tt mawu} $\mid$
				$\varepsilon$ $\mid$
				{\tt ma} 
				
\end{compactenum}
and extending the grammar rule for {\sl v} with the extra option:

 {\sl v} $\rightarrow$ {\sl pre vr post} {\tt a} $\mid$
						{\sl npre vr post} {\tt i} $\mid$
						{\sl ppre vr} {\tt e}

\paragraph{Stative verbs} The stative refers to the state of being of something; e.g. {\em vula} (`open') with its stative variant {\em vuleka} (`be opened'), and {\em mbula} (`reveal')  results in {\em mbuleka} (`be revealed'). This insertion of the {\em -ek-} between the VR and the FV is also referred to as the neuter extension. As it is conceptually different from the extension (i.e., CARP), we create a separate {\sl st} and update the rule for {\sl v} with it:

 {\sl v} $\rightarrow$ {\sl pre vr post} {\tt a} $\mid$
						{\sl npre vr post} {\tt i} $\mid$
						{\sl ppre vr} {\tt e} $\mid$ {\sl vr st} {\tt a}

\noindent with the following single terminal:
\begin{compactenum}
\setcounter{enumi}{11}
\item Stative verb, indicated with {\sl st}:

{\sl st} $\rightarrow$ {\tt ek} 

\end{compactenum}
Because there is only one terminal for {\sl st}, the ``{\sl vr st} {\tt a}''-part of {\sl v} can also be written as ``{\sl vr} {\tt eka}''.

\paragraph{Wh-questions} The optional wh-questions fall in the post-final slot (see Table~\ref{tab:slots}) and are added at the end of the verb, being {\em -ni}  `what'/`who'/`why'/`how', {\em -nini} `when', and {\em -phi} `where'. We create a separate {\sl wh} variable for them and update the rule for {\sl v} with it:

{\sl v} $\rightarrow$ {\sl pre vr post} {\tt a} {\sl wh} $\mid$
						{\sl npre vr post} {\tt i} {\sl wh} $\mid$
						{\sl ppre vr} {\tt e} $\mid$ 
						{\sl vr st} {\tt a} 

\noindent with the following terminals for the new variable:
\begin{compactenum}
\setcounter{enumi}{12}
\item Wh-questions, indicated with {\sl wh}:

{\sl wh} $\rightarrow$ {\tt ni} $\mid$ {\tt nini} $\mid$ {\tt phi} $\mid$ $\varepsilon$ 

\end{compactenum}

\paragraph{Aspect doubling} What is normally referred to as aspect doubling is a construction of aspect with continuous tense, i.e., the `second aspect' is not an aspect in the strict sense of the meaning of aspect. Decomposed, we have EXCL-SC-CONT-(OC-)VR-(post)-a, where the exclusive can only be {\em se-} and continuous tense only {\em -ya-}. Because it is a regular exception, we add another `or' to {\sl v} rather than complicate {\sl pre}:

{\sl v} $\rightarrow$ {\sl pre vr post} {\tt a} {\sl wh} $\mid$
						{\sl npre vr post} {\tt i} {\sl wh} $\mid$
						{\sl ppre vr} {\tt e} $\mid$ 
						{\sl vr st} {\tt a} $\mid$ \\
\indent\indent\indent						{\sl excl s cont o vr post} {\tt a}

\noindent with the following terminals for the new variables:
\begin{compactenum}
\setcounter{enumi}{13}
\item `Double aspect', indicated with {\sl excl} for exclusive (with  {\sl excl $\subset$ asp}) 

{\sl excl} $\rightarrow$ {\tt se} 

\item With {\sl cont $\subset$ t} and {\sl cont}  for continuous tense:

{\sl cont} $\rightarrow$ {\tt ya} 

\item The previous extension implies that {\sl t} (item 3, above) also has to be updated:

{\sl t} $\rightarrow$ {\tt ya} $\mid$ $\varepsilon$

\end{compactenum}
Finally, there is only one terminal for each, so the ``{\sl excl s cont o vr post} {\tt a}''-part of {\sl v} can also be written as ``{\tt se} {\sl s} {\tt ya} {\sl o vr post} {\tt a}''.

\subsection{Other rules}
\label{sec:otherrules} 
While the CFG may seem alike a relatively free combination of anything, there are several constraints that are not covered by these grammar rules, as they would obfuscate the general patterns, not all of them are linguistically accounted for, and they are easier to implement as separate rules. Notably, there is an interaction between the two sides of the VR, which is ruled by the semantics of the CARP extension. For instance, for a construction to be causative and applicative, there have to be at least two things involved. The first participant is already catered for with the SC, the second is catered for with the OC. Typically, the causative and applicative will have an OC but the Reciprocal, Passive and the Stative would not. Further, for the passive, the object moves to the subject position, and so also with SC and OC. 

The following set of rules (in pseudocode-style notation) is a first attempt at specifying them, and more will be added in due course:
\begin{compactenum}[a)]
\item only {\bf if} {\sl p $\in$ post}, {\bf then}: {\bf if} {\sl pre} {\bf then} {\sl s} $\rightarrow$ $\varepsilon$, {\bf else} {\sl ns} $\rightarrow$ $\varepsilon$  
\item {\bf if} {\sl c $\in$ post}, {\bf then} {\sl s,o $\in$ pre} or {\sl npre}
\item {\bf if} {\sl a $\in$ post}, {\bf then} {\sl s,o $\in$ pre} or {\sl npre}
\item {\bf if} {\sl p $\in$ post}, {\bf then} {\sl o} $\in$ ({\sl pre} or {\sl npre}) and {\sl s} $\notin$ ({\sl pre} or {\sl npre}) 
\item {\bf if} {\sl vr} $\in$ {\sf Intransitive}, {\bf then} {\sl r} $\notin$ {\sl post} and {\sl o $\notin$} ({\sl pre} or {\sl npre})
\item {\bf if} {\sl vr} $\in$ {\sf Monosyllabic}, {\bf then} {\sl post} $\rightarrow$ $\varepsilon$
\end{compactenum}

\noindent The second set of rules have to do with phonological and morphological conditioning, such as:
\begin{compactenum}[a)]
\setcounter{enumi}{6}
\item {\bf if} {\sl s}=={\tt u} {\bf then} {\sl pl}={\tt aw}, {\bf else} {\sl pl}={\tt aw} or {\tt mawu} or {\tt ma}
\end{compactenum}
which we consider orthogonal to coverage of the different elements of the verb, and is therefore left for further work.

\subsection{Extensions and other considerations}
\label{sec:disc}

While the `other rules' indicate intricate interactions between the various elements of a verb that might be addressed either with extra-CFG rules or a blow-up of CFG rules (by splitting the current ones in various ways) once fully known, one of a different kind is the treatment of the elements themselves. For instance, TAM can be at the start of the verb and at the end, but when at the start or end, only a {\em subset} of TAM is permissible, i.e., FV $\subset$ ? $\subset$ TAM, where that subset ``?'' exists, but is not well-documented yet as to why, what, and how. 

The formal approach taken in the previous section lends itself well to a rigorous assessment of measurable distance or difference with verbs in other Bantu languages, as well as bootstrapping a CFG for some of the closely related but even lesser-resourced languages, such as Ndebele and isiXhosa. That said, we are well aware it will not be the same. Take, for instance, the Chishona---a neighbouring language---example in V5. 

$
\begin{array}{llll}
\mbox{(V5)} & \mbox{\em mukomana} &	\mbox{\em a-ri-ku-gur-ir-a-zve} 	& \mbox{\em chisikana}\\
		& \mbox{1.boy} & \mbox{1.SC-T-M-break-APPL-FV-too} & \mbox{7.girl}\\
\end{array}
$

	\hspace{8mm} `The boy is breaking (something) for the girl too'

\noindent The order of the extensions and clitics in the example above is worth noting. The clitic {\em -zve} comes after the FV {\em -a}. While this phenomenon is not found in isiZulu verb complex, it shows that there are unique features of the Bantu verb that further complicate the grammar, which will need to be accounted for.  

Finally, incorporation of phonological and morphological conditioning, while being an orthogonal aspect to the structure of the components of the verb and order thereof, may, for practical reasons, have an effect on the rules itself. For instance, possibly splitting the terminals of the {\sl vr} into one set for vowel-commencing roots and one for consonant-commencing roots, and then for ease of processing, some of the terminals of the {\sl pre} and {\sl npre} could be split into two as well.

\section{Evaluation of the grammar}
\label{sec:eval}

We first illustrate manually the functioning of the CFG with three use cases, and subsequently test it systematically with a computational version of it.

\subsection{Use cases}

Three examples are selected that also give a hint toward the CFG's usability for a range of applications: generation of a word from the grammar (useful for machine translation and controlled natural languages), the checking of a correct word whether it is in the language of the grammar (spellchecking), and one misspelled word that gets rejected (correcting). 
				
Let us step through the grammar in the least amount of steps (least amount of components) to `generate' a word in the language, where each numbered subscript of the arrow is added for explanatory purpose afterward: 
{\sl v} $\Rightarrow_1$ {\sl pre vr post} {\tt a} {\sl wh} $\Rightarrow_2$ {\sl s vr post} {\tt a} {\sl wh} $\Rightarrow_3$ {\tt ngi} {\sl vr post} {\tt a} {\sl wh} $\Rightarrow_4$ {\tt ngi vel} {\sl post} {\tt a} {\sl wh} $\Rightarrow_5$ {\tt ngi vel a} {\sl wh} $\Rightarrow_6$ {\tt ngi vel a}; 
1) substitute {\sl v} for the first option; 2) substitute {\sl pre} for the first option ({\sl s}); 3) substitute {\sl s} with the first terminal ({\tt ngi}); 4) take one of the {\sl vr}s ({\tt vel}); 5) process {\sl post}, choosing empty ($\varepsilon$); 5) process {\sl wh}, choosing empty ($\varepsilon$). Thus, the word generated is: {\tt ngivela} `I come from'. 
%

Stepping through the grammar, using more slots at the end, we can check that {\tt niboniselana} is in the language:
{\sl v} $\Rightarrow$ {\sl pre vr post} {\tt a} {\sl wh} $\Rightarrow$ {\tt ni} {\sl vr post} {\tt a} {\sl wh} $\Rightarrow$ {\tt ni bon} {\sl post} {\tt a} {\sl wh} $\Rightarrow$ {\tt ni bon} {\sl c a r} {\tt a} {\sl wh} $\Rightarrow$ {\tt ni bon is} {\sl a r} {\tt a} {\sl wh} $\Rightarrow$ {\tt ni bon is el} {\sl r} {\tt a} {\sl wh} $\Rightarrow$ {\tt ni bon is el an} {\tt a} {\sl wh} $\Rightarrow$ {\tt ni bon is el an} {\tt a}. 

The grammar thus also can be used to recognising misspelled words. For instance, a user types $^*${\tt usafundapi}, then it rejects at the {\tt -pi} end: {\sl v} $\Rightarrow$ {\sl pre vr post} {\tt a} $\Rightarrow$ {\sl s asp vr post} {\tt a} {\sl wh} $\Rightarrow$ {\tt u} {\sl asp vr post} {\tt a} {\sl wh} $\Rightarrow$ {\tt u sa} {\sl vr post} {\tt a} {\sl wh} $\Rightarrow$ {\tt u sa fund} {\sl post} {\tt a} {\sl wh} $\Rightarrow$ {\tt u sa fund a} {\sl wh} $\Rightarrow$ $\times$. The trace/tree can not be completed because {\tt pi} $\notin$ {\sl wh} ({\tt phi} and {\tt ni} are), thus $^*${\tt usafundapi} is misspelled with respect to the grammar rules as introduced in the previous sections. 
Proposing a correction can be done by suggesting to complete {\tt usafunda}- with any of the {\sl wh} terminals, or, when using the minimum edit distance as an extra service in the spellchecker, it would suggest {\tt usafundaphi} `where are you still studying?' and {\tt usafundani} `what/why are you still studying?' as the two options to choose from to correct the misspelled word.

\subsection{Computational evaluation of the grammar}
\label{sec:algs}

There are many tools that are candidates to implement the grammar to the point of testing whether the rules are the right ones; that is, the scope is {\em validation} (`are we building the right grammar?') and {\em verification} (`are we building the grammar right?'), not end-user tool building.

\subsubsection{Implementation considerations}

Most computational linguistics papers for Bantu languages use one of the tools for building a morphological analyser. Xfst and lexc has been used to encode a subset of the rules for verbs in isiZulu, Setswana, and  Ekegusii \cite{Pretorius03fsm,PretoriousR09,Elwell05}, whereas Fsm2 could not be found online anymore. However, they are problematic theoretically. Xfst, and similar tools such as SFST\footnote{\url{http://www.cis.uni-muenchen.de/~schmid/tools/SFST/}} and OpenFST\footnote{\url{http://www.openfst.org/twiki/bin/view/FST/WebHome}}, are transducers, and therewith limited to regular grammars (The surface syntax gives the impression of accepting a CFG, but that is syntactic sugar and is transformed behind-the-scenes into a (very) large FSA). While most of the rules in the previous section look indeed regular, for being in a fixed order at least, when {\sl p} $\in$ {\sl post}, then the {\sl o} $\in$ {\sl pre} takes the position of the {\sl s}. This already indicates that the grammar for the verb on its own is beyond a regular grammar, hence, beyond a FSM, so a transducer is insufficiently expressive. This is unsurprising, as natural languages tend all to be context-free \cite{Pullum82}. Another option is to take a programmatic approach. Python programming language is popular, used by \cite{Spiegler10}, and the NLTK \cite{Bird09} has a CFG grammar module. However, the latter requires the word already to be segmented, but this is precisely what needs to happen automatically, and building a regular expression grammar faces the same issue as mentioned above. Spiegler et al's DCG for the Ukwabelana tagging is in Prolog, but at this validation and verification stage a full-fledged tool is not needed. Therefore, we used the JFlap tool\footnote{\url{http://www.cs.duke.edu/csed/jflap/}}, which can check string membership and generate words in the language.

\subsubsection{Testing in JFlap}

Transferring the written grammar into JFlap (v8 beta) ironed out two glitches in variable abbreviations (corrected version is included in the previous section), and some of the variable names are different, because the tool allows only single-character variable names. The JFlap file, conversion annotations, and the screenshots of the outputs are available online at \url{http://www.meteck.org/files/geni/}.

\begin{figure}[h]
\centering
  \includegraphics[width=0.6\textwidth]{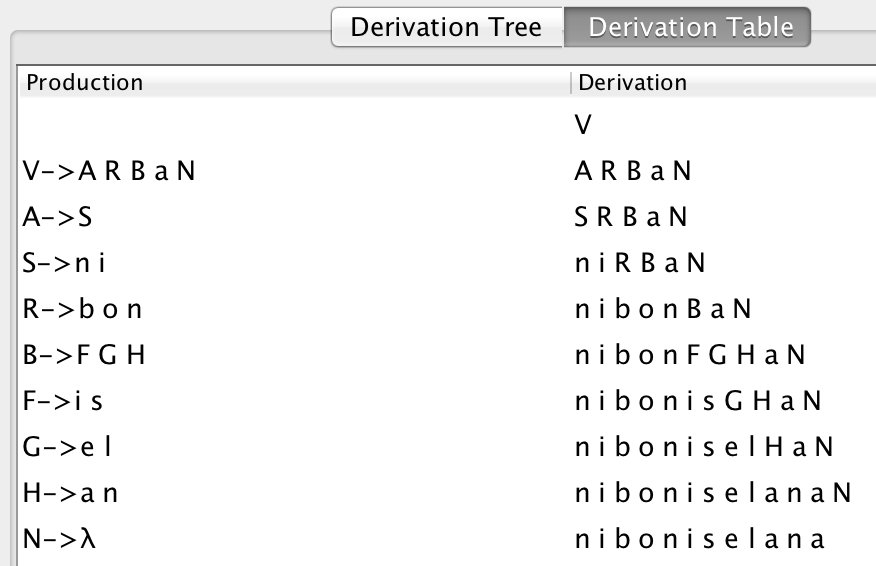}
\caption{Screenshot of the derivation table as computed by JFlap (brute force parser), on {\tt niboniselana}.}
\label{fig:flapss2}       
\end{figure}

We selected a set of verbs that covers the principal permutations of the rules, and some that ought to be rejected, as indicated in Table~\ref{tab:jflaptest}. 
The strings in the first set were all accepted; a screenshot of the derivation table of {\tt niboniselana} is shown in Fig.~\ref{fig:flapss2} and the screenshots for the others are in the online material. 
Thus, the CFG recognises what it should recognise, and thus indicates correctness of the grammar specified. 
Of the terms one would have liked to have it rejected, only {\tt ngiveli} was (incorrectly) accepted,
which is due to the $\varepsilon$'s that are in the grammar due to the absence of the extra rules in the JFlap CFG (recall Section~\ref{sec:otherrules}), so {\sl npre} is decomposed as {\sl ns o}, with {\sl ns} $\Rightarrow \varepsilon$ and {\sl o} $\Rightarrow$ {\tt ngi}. Thus, the  strings that the grammar accepts/generates are more words than that are in the isiZulu language. This can be seen also with the `generate strings' feature in JFLap. Setting the number of strings to a subset, 100, to check, showed that this is due to not only not including the additional constraints but also not catering yet for phonological conditioning; e.g., the aforementioned $^*${\tt ngiveli} and $^*${\tt aabaphi}, where {\tt a}=SC nc:6, {\tt a} = Mood, {\tt ba} = VR `distribute', and {\tt phi} = wh `where', but it requires an consonant between the two a's. Addressing this issue is orthogonal to the grammar, and therefore left for future work.

\begin{table}[h]
\caption{Strings selected for testing; A/R: accepted/rejected.}
\begin{center}
\begin{tabular}{|p{2.6cm}|l|c|c|}
\hline
{\bf String} & {\bf Reason} & {\bf A/R} & {\bf Correct} \\  \hline   \hline           
 {\tt ngivela} & simple present tense & A & + \\  \hline
{\tt angiveli} & simple negation & A & + \\  \hline
{\tt angivela} &  {\sl pre} with {\sl s} and {\sl o} & A & + \\  \hline
{\tt asingabaveli} & testing {\sl npre} & A & + \\  \hline
{\tt niboniselana} & testing {\sl post} & A & + \\  \hline
{\tt vuleka} & stative & A & + \\  \hline
{\tt usafundaphi} & wh-extension & A & + \\  \hline
{\tt sengiyafunda} & aspect doubling & A & + \\  \hline
{\tt awusidle} & politeness & A & + \\  \hline \hline
$^*${\tt ngiveli} & mixing positive with negative FV & A & -- \\  \hline
$^*${\tt usafundapi} & typo; wrong wh-extension & R & +  \\  \hline
$^*${\tt nibonelisana} & wrong CARP order & R & + \\  \hline
$^*${\tt sangiyafunda} & wrong aspect in aspect doubling & R & + \\  \hline
 $^*${\tt kabevela} & wrong order in {\sl pre}, no {\sl sc} & R & + \\ \hline
\end{tabular}
\end{center}
\label{tab:jflaptest}
\end{table}%

Further, we checked our grammar against the Zulu finite state morphology demo of the Academy of African Languages and Science\footnote{\url{http://gama.unisa.ac.za/demo/demo/zulmorph}; tested with the version online d.d. 17-12-2015.}. As one may expect, it accepts {\tt niboniselana}, but it also accepts $^*${\tt nibonelisana}, which has the wrong CARP order, and accepts $^*${\tt kabevela} (MOD-ASP-VR-FV), which has MOD and ASP in the wrong order (and lacks SC/OC) and is therefore rejected by our CFG. That is, that FSM does not handle any order of components of the verb.

\begin{figure}[h]
\centering
  \includegraphics[width=0.9\textwidth]{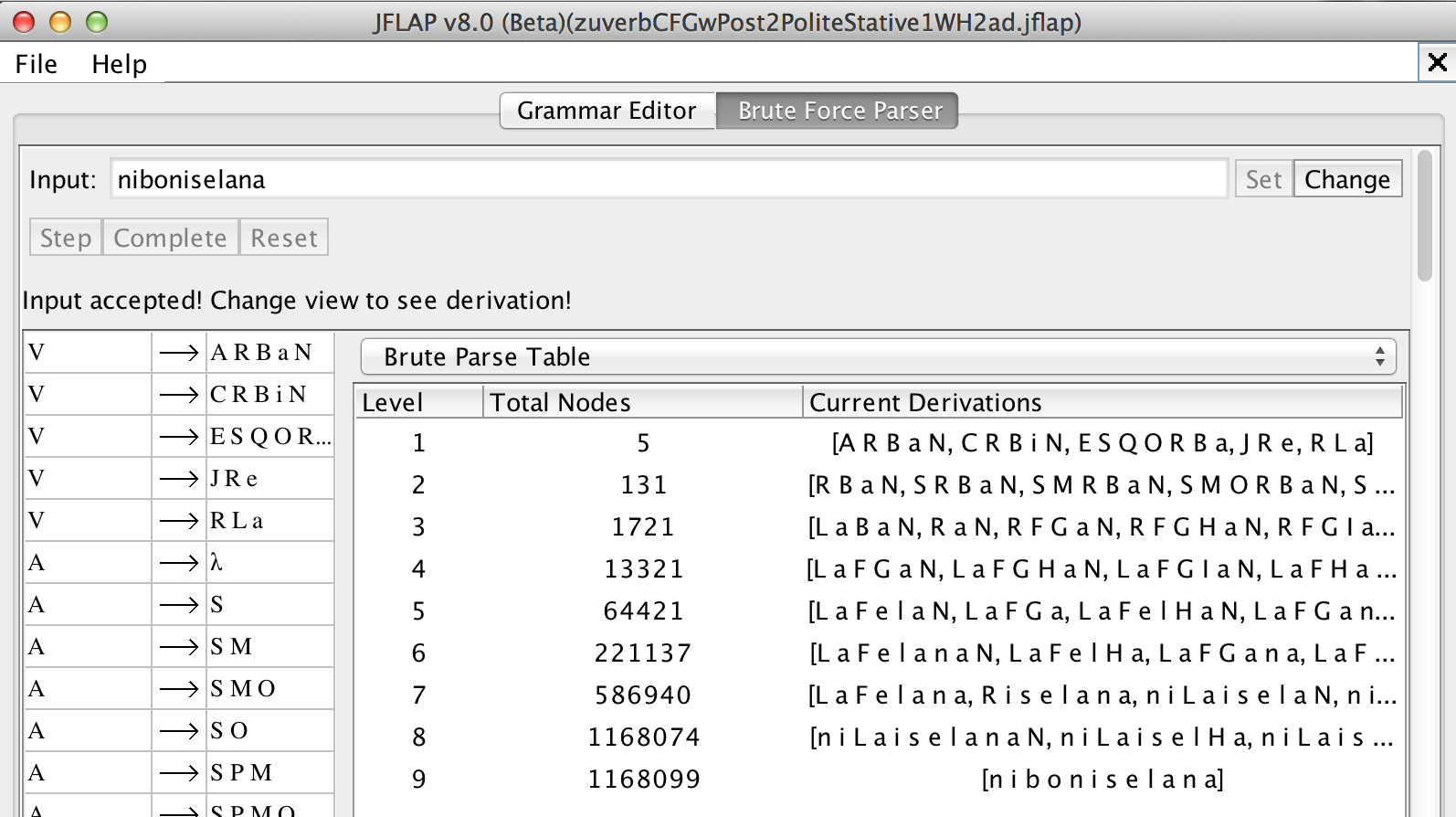}
\caption{Screenshot of JFlap's output with the brute force parser for the final CFG, on {\tt niboniselana}.}
\label{fig:flapss}       
\end{figure}

While using a CFG computationally is an error-proof method for `finding' a derivation, the CFG up to the wh-extension used up 773240 nodes in the brute force parser to accept {\tt niboniselana}, due to exploring all potential paths to completion. This was with five verb roots for testing the grammar; adding another five verb roots generated 1086109 nodes in order to accept {\tt niboniselana}. With the aspect doubling extension, this increased further to 1168099 nodes; see Figure~\ref{fig:flapss}. Computationally, this is clearly not sustainable with a brute force parser, and a practical implementation that uses the CYK algorithm instead will be needed. Nevertheless, the brute-force parser is useful for evaluating the grammar.

\section{Conclusions}
\label{sec:concl}

We presented the precise specification of the isiZulu verb present tense as a Context-Free Grammar. It covers not only the usual subject and object concords, but also negation, present tense, aspect, mood, and the verbal extensions such as the causative, applicative, stative and the reciprocal, politeness, the wh-questions modifiers, and aspect doubling, all in their correct order as they appear in verbs. In addition to a paper-based specification, it was represented computationally as a CFG in the JFlap tool and tested on correctness of specification, using a set of words and generating their derivations in the JFlap tool. The grammar conforms to specification, though still accepts more strings than those that are in the isiZulu language. This is due to the absence of additional rules in the implementation and the orthogonal issue of phonological conditioning, which are aspects of future work.

\subsubsection*{Acknowledgements}
This work is based on the research supported in part by the National Research Foundation of South Africa (CMK: Grant Number 93397).


%


\end{document}